\documentclass[letterpaper, 10 pt, conference]{ieeeconf}  

\IEEEoverridecommandlockouts                              

\usepackage{amsmath} 

\usepackage{gensymb}
\usepackage[mathscr]{euscript}
\usepackage{graphicx}
\usepackage[labelformat=simple]{subcaption}

\title{\LARGE \bf
Deliberate Exploration Supports Navigation in Unfamiliar Worlds
}

\author{Raj Korpan$^{1}$ and Susan L. Epstein$^{2}$
\thanks{*This work was supported in part by The National Science Foundation under IIS \#162584}
\thanks{$^{1}$Raj Korpan is with the Department of Computer Science of
        The City University of New York, New York, NY 10016, USA
        {\tt\small rkorpan@gradcenter.cuny.edu}}%
\thanks{$^{2}$Susan L. Epstein is with the Department of Computer Science, Hunter College and The Graduate Center of
        The City University of New York, New York, NY 10065, USA
        {\tt\small susan.epstein@hunter.cuny.edu}}%
}

\begin{document}

\maketitle
\thispagestyle{empty}
\pagestyle{empty}

\begin{abstract}
To perform tasks well in a new domain, one must first know something about it. This paper reports on a robot controller for navigation through unfamiliar indoor worlds. Based on spatial affordances, it integrates planning with reactive heuristics. Before it addresses specific targets, however, the system deliberately explores for high-level connectivity and captures that data in a cognitive spatial model. Despite limited exploration time, planning in the resultant model is faster and better supports successful travel in a challenging, realistic space.

\end{abstract}

\section{Introduction}
A \textit{cognitive spatial model} is a mental representation that captures connectivity, distances, patterns, and topological structure \cite{RN2351}. Navigators learn such a model based on their perception and their experience \cite{golledge1999human}. The thesis of our work is that learned generalizations over spatial data can efficiently and effectively support robot navigation without, or in addition to, an explicit metric map. This paper reports two key advances toward that goal. First, deliberate exploration for high-level connectivity develops a more complete and robust spatial model. Second, path planning with such a model is more effective and efficient. Empirical results reported here demonstrate the impact of our approach on a simulated robot without a metric map in a challenging, realistic world. 

\begin{figure}[t]
\centering
\includegraphics[width = 0.94\linewidth]{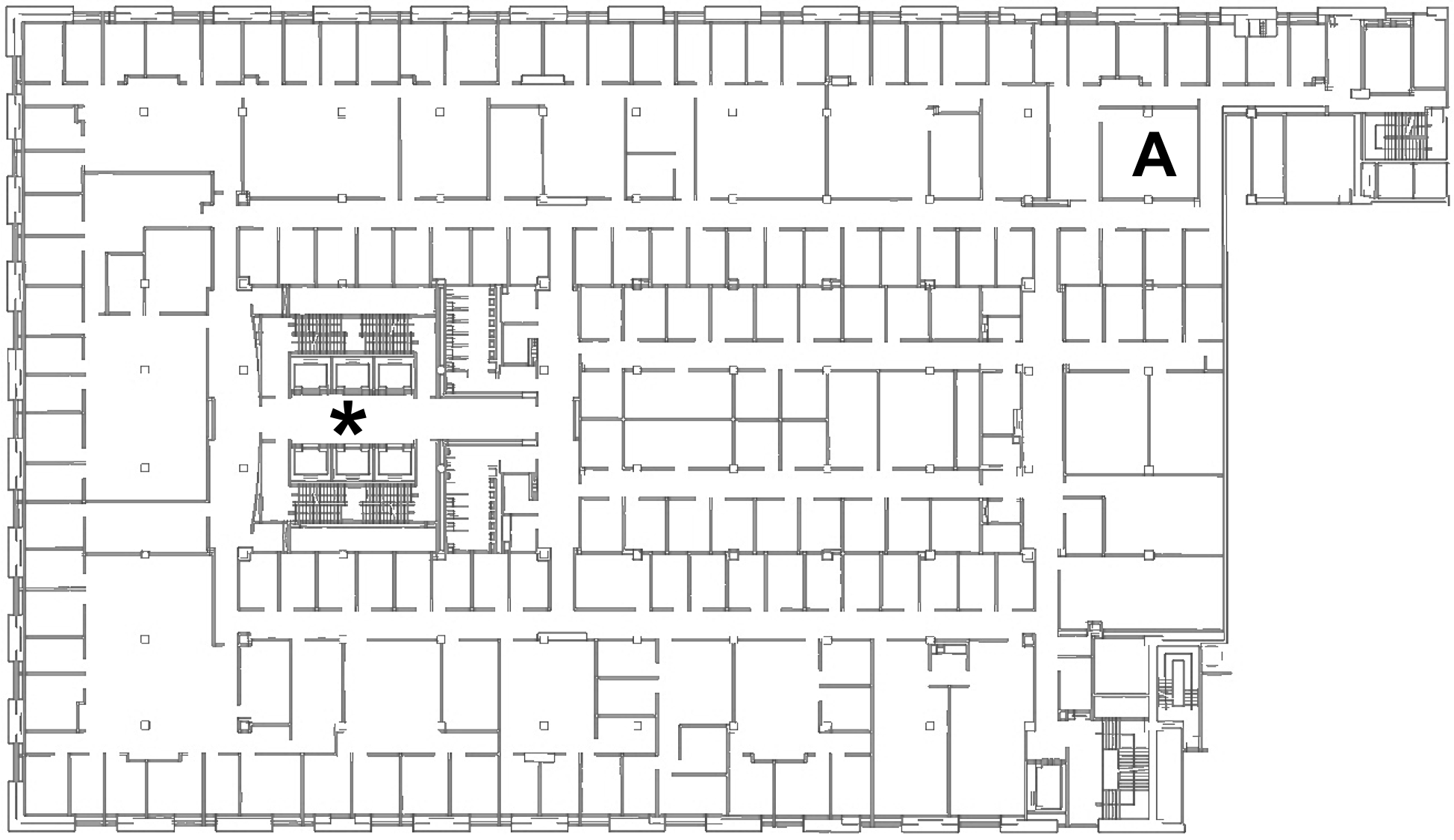}
\caption{How would you travel in GCWorld from * to A?} 
\label{fig:GCWorld}
\end{figure}

Imagine that you must travel from the elevators at * to the room labeled A in \textit{GCWorld}, shown in Figure \ref{fig:GCWorld}. Look carefully at the map. How would you get there? What would simplify your task? If you recognize and plan with the long hallways in Figure \ref{fig:GCWorld}, the task becomes far easier. GCWorld is the fifth floor of a $110m \times 70m$ office building that occupies an entire city block. An architect presumably intended its hallways to facilitate travel. They extend, however, beyond the $25m$ range of many robots' laser range finder.

\begin{figure}[b]
\centering
\includegraphics[width = 0.4\linewidth]{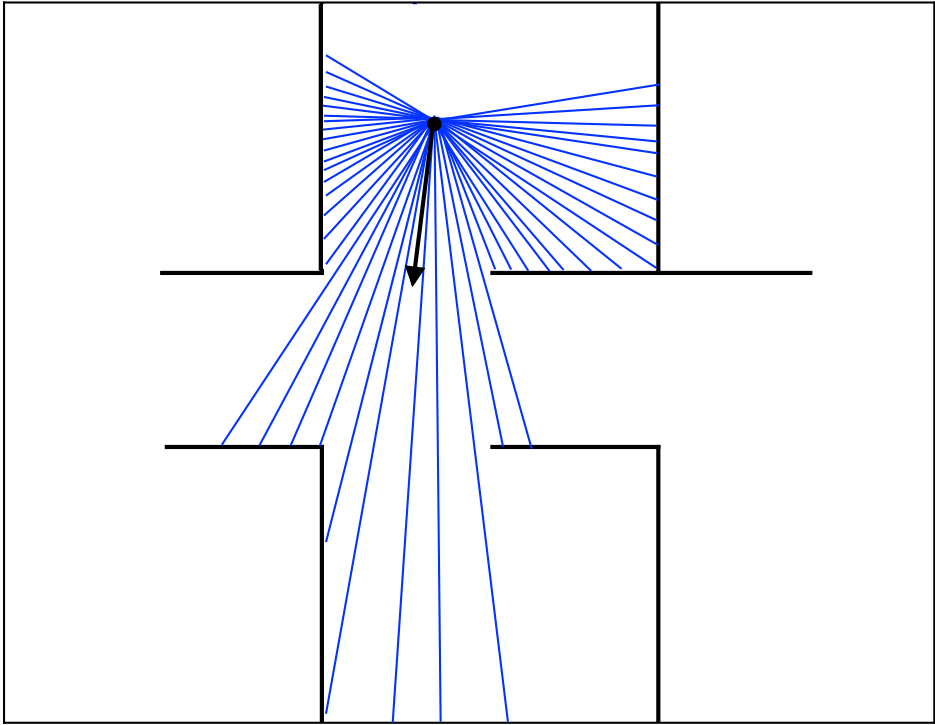}
\caption{A robot's view} 
\label{fig:laserscan}
\end{figure}

\textit{SemaFORR} is a robot controller, written in ROS, to address space as supportive rather than restrictive. Like human navigators, SemaFORR learns \textit{spatial affordances}, features of the world that enable, not hinder, navigation. The robot's \textit{view} in any location is the raw data reported by its laser range finder. As in Figure \ref{fig:laserscan}, a view reports finitely-many distances to the nearest obstruction, measured along an arc for peripheral vision. A robot's \textit{pose} (location and orientation with respect to some fixed point) has a unique view.

In earlier work, SemaFORR learned local spatial affordances that arose felicitously as it traveled to targets \cite{epstein2019planning}. Combined with planning in a detailed metric map, these affordances improved SemaFORR's ability to contend with large, novel, irregular spaces. Here, however, we assume the robot has no map, and instead have it actively seek out more global affordances beyond its immediate perceptual range. 

A vehicle highway system is in some ways analogous to GCWorld's network of hallways. A plan to travel to a specified destination through a connected system of highways requires an initial path to the system itself, travel along some portion of it, and a final path to the destination. Highways, like hallways, provide long-distance segments with high-level connectivity. Highways' restricted on-off access, however, reduces opportunities for inefficient digressions. In contrast, a navigator in GCWorld faces unknown possibilities at every doorway. Highway systems are intended to facilitate travel; we assume here that architects design buildings with the same intent. Thus SemaFORR treats hallways the way navigation apps treat highways --- it plans with them. But to use such a network, SemaFORR must first discover it. 

The next section describes related work on exploration and navigation. Subsequent sections detail new algorithms that learn high-level connectivity by deliberate exploration, and build a cognitive spatial model to navigate an unfamiliar world. The paper then describes our navigation experiments, analyzes their results, and discusses current and future work.
 
\section{Related Work}
To travel in an unknown world, a navigator explores, represents observed space, and then plans with that representation. One such representation is a metric map that accurately positions every obstacle. The state-of-the-art algorithm to build a detailed metric map of the world for robot navigation is Probabilistic SLAM (Simultaneous Localization and Mapping) \cite{montemerlo2007fastslam}. To produce end-to-end plans, however, SLAM-based navigation typically requires complete coverage of a world. This requires painstaking travel along each wall, particularly burdensome in large, complex spaces like GCWorld with its jogs and connecting rooms. SLAM must also contend with loop-closing and it treats all space as equally important.

Several approaches to a metric map prioritize exploration. Frontier-based exploration repeatedly moves to the closest unknown space while any remains \cite{yamauchi1997frontier}. This can be inefficient. Recent work used context and the location of a goal to bias frontier-based exploration \cite{everett2019planning}. In contrast, our approach prioritizes exploration of space with high connectivity for navigation to any goal.

Instead of a metric map, many human navigators formulate a cognitive spatial model based on prior knowledge, external cues, and sensory input (e.g., vision and proprioception). Place cells and grid cells in the human hippocampus construct a spatial model \cite{bellmund2018navigating}. People learn a labeled graph that captures topological information with metric labels \cite{warren2017wormholes}. With deliberate but relatively limited exploration, SemaFORR also learns a labeled graph that captures the highly connected, most salient spatial relations in the robot's world.

Several hierarchical representations of the world that capture connectivity require the robot to navigate through every part of its world \cite{beeson2010factoring}. People, in contrast, do not require a complete map or physical traversal along every wall to navigate well. Instead, they look for highly-connected thoroughfares that facilitate efficient travel \cite{holscher2007capturing}. Our approach enables hierarchical planning at different levels of abstraction, similar to the way a human brain constructs, represents, and uses hierarchical plans \cite{balaguer2016neural}. 

Without a spatial representation, reinforcement learning for navigation often struggles with large, realistic worlds \cite{savva2017minos}. Although one recent approach achieved near perfect success this way, it learned from billions of simulations of robots equipped with three sensors \cite{wijmans2019dd}. In contrast, our approach learns to navigate successfully with limited exploration, a range finder, and a cognitive spatial model.

Classical graph planning algorithms (e.g., breadth-first search and A* \cite{korf2014search}) construct a graph and search it for shortest paths. Such graphs can be very large because they locate many vertices in every spatial area. Instead, sampling-based methods (e.g., probabilistic roadmap and rapidly-exploring random tree (RRT)) create smaller graphs by randomly sampling free space. Such graphs may not include the shortest paths, however. While other sampling-based methods (e.g., RRT* \cite{karaman2011sampling} and DeRRT* \cite{kuo2018deep}) find near optimal paths, they still require a spatial representation of the entire world, and their paths can come close to obstacles. Our approach uses a classical planning algorithm on a labeled graph whose vertices avoid obstacles in explored areas instead of a graph based on a metric map.

An \textit{affordance} is a characteristic of the world that enables the execution of some action \cite{gibson1977theory}. Affordance-based theories of spatial cognition posit a tight relationship between the specific dynamics of the environment and the decisions made by an individual \cite{fajen2013spatial}. To the best of our knowledge, only one system other than SemaFORR has addressed navigation and spatial representation hierarchically and with affordances \cite{pronobis2017deep}. The space it explored, however, was an order of magnitude smaller and hardly more complex than Figure \ref{fig:laserscan}. Its model also required world-specific deep learning offline and designated semantic labels for room types. In contrast, SemaFORR produces a cognitive spatial model based on spatial affordances, and learns online, both from deliberate exploration and task-driven experience.

\section{Exploration for High-Level Connectivity}
\begin{figure}[t]
\centering
\includegraphics[width = \linewidth]{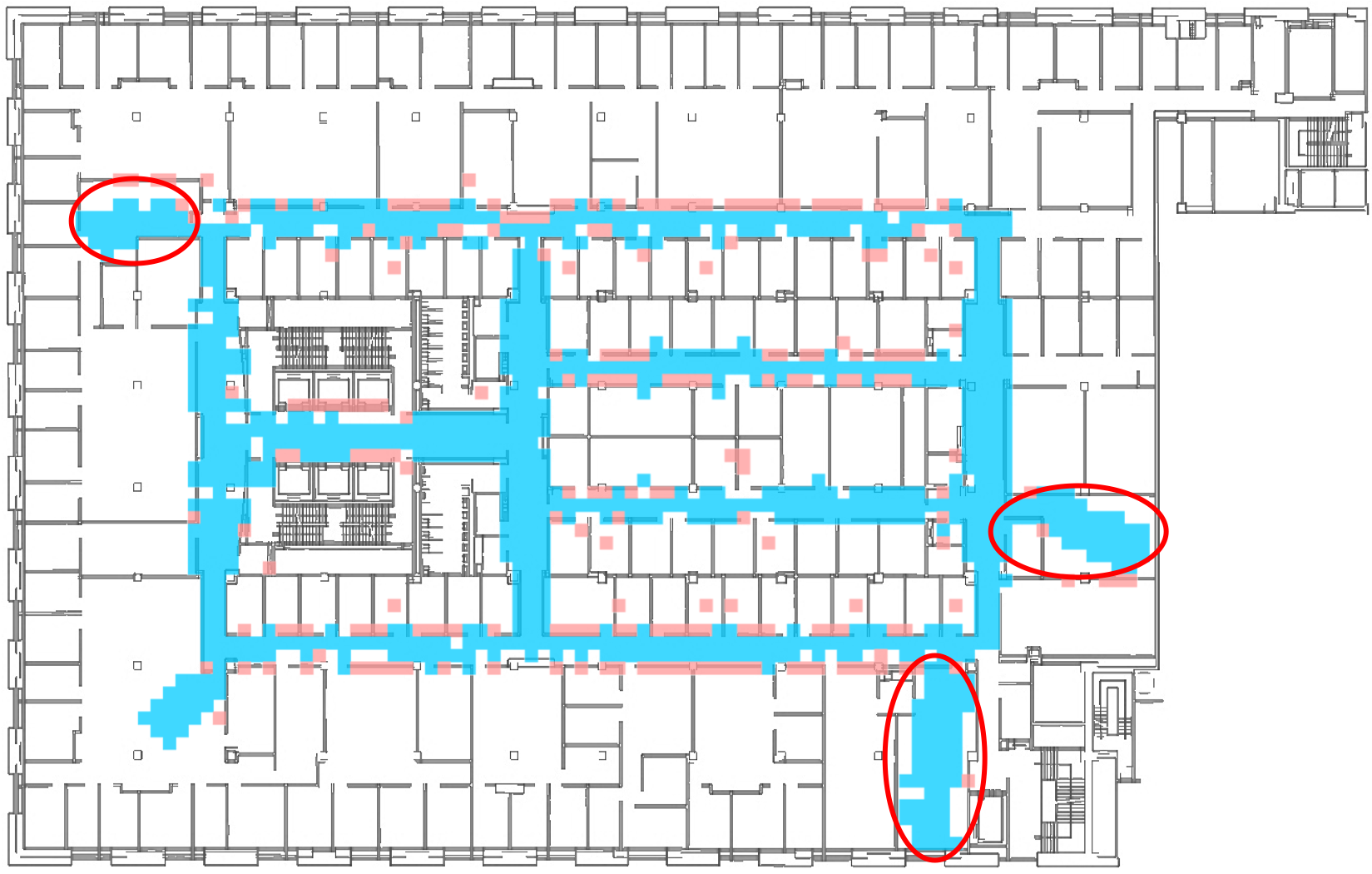}
\caption{Passages (blue) and obstructed cells (pink) learned after 20 minutes of exploration. Rooms that eluded the termination condition are circled.} 
\label{fig:highwayexample}
\end{figure}

Faced with a novel task, people often prefer to explore for relevant global information, rather than take the most immediately rewarding action \cite{collignonepistemic}. This section details \textit{HLC} (High Level Connectivity), a one-time, view-based process to explore built, indoor spaces. Figure \ref{fig:highwayexample} resulted from 20 minutes of exploration with HLC in GCWorld. This exploration is heuristic, opportunistic, and time-limited.

Confronted with a new built world, HLC seeks \textit{passages}, relatively far-reaching, relatively narrow, relatively regular extents intended to approximate hallways. Based solely on the robot's views, HLC identifies likely passages (\textit{candidates}) for exploration. The remainder of this section describes how HLC identifies and stores candidates, carefully examines one at a time, and reformulates its knowledge into a graph that supports planning. 

\subsection{Finding candidates}
In HLC, exploration is a sequence of decision points. At each decision point, HLC selects and executes one action. Allowable actions are forward moves of six different lengths and four sizes of rotation, both clockwise and counterclockwise. For every decision point, HLC records the robot's pose and view, along with the features in Table \ref{tab:features}. 

Within a view, a potential passage first appears to the robot as a \textit{stretch}, a long, thin, unobstructed area of length at least $d$. A candidate is a stretch and its \textit{start}, the pose where it was detected. Initially, HLC rotates the robot in place to find its first candidate. Although HLC pursues only one candidate at a time, the robot may glimpse other stretches as it pursues its current candidate. HLC saves qualified stretches for future exploration on its \textit{candidate list}. A candidate is placed at the front of the list if its average length is more than $2*d$; otherwise it is placed at the back. This prioritizes exploration of the largest, most recently glimpsed stretches first. 

To qualify as a candidate, a stretch and the pose where it is detected must satisfy four conditions. First, the stretch's length must exceed its width. Second, the stretch is not similar to any candidate on the list or any previously considered one. (Two stretches are judged similar if the distance between them is within $1m$ and some Allen interval condition applies \cite{allen1983maintaining} with an overlap of at least $1/3$ the average of their lengths.) Third, the stretch must potentially provide new, useful information. To determine this, HLC keeps a \textit{passage grid} of \(1m\times 1m\) cells superimposed on the footprint of the world. Each cell there can be labeled as part of a passage, labeled as obstructed (by a process described below), or left unlabeled. HLC gauges the potential for new, useful information based on the start, midpoint and end of the stretch. The third condition is that none of those points is labeled as obstructed in the grid, and no more than one is labeled as part of a passage. The fourth condition is that the candidate may not be classified as a room.

Large rooms can be mistaken for a stretch when they are actually dead ends. To prevent their candidacy, the view statistics in Table 1 were used to distinguish between a large room and a legitimate continuation of a passage. Offline, a large set of unlabelled views was drawn from earlier experiments, and described by the features in Table 1. We clustered this data with $k$-means into two sets, which had natural labels as rooms and passages in the world. We labeled each data point by its cluster, learned a decision tree on this labeled data, and made its top two rules a classifier that distinguished between a passage and a room. 

\subsection{From candidate to passage}
To explore a candidate, HLC assigns it a passage number and then navigates to reach the farthest position along its stretch. As it moves the robot, HLC records the passage number at each grid cell it enters. Real hallways, however, are not necessarily smooth; they often have minor jogs and slight irregularities. If the robot comes too close to an obstruction on its left or right while it traverses the stretch, HLC has it turn slightly away from the obstruction, take a step, and then continue down the stretch. 

Of course, a passage may be longer than a candidate's stretch initially suggests. For example, the passage might curve slightly or exceed the range finder's limit. To contend with these possibilities, exploration of a candidate stops only when one of three conditions is satisfied. First, the robot has reached the end of the passage (i.e., lies within $0.5 m$ of the end of the stretch or has only $0.1 m$ directly in front of it). Second, the robot has just made so hard a turn that it would pursue a different passage (i.e., its current orientation differs by more than $45 ^\circ$ from its average orientation in the current passage thus far). Third, the robot may find itself in a large space or room, detected when the space around the robot noticeably widens (i.e., the current passage's length plus the maximum visible distance in front of the robot is less than 1.5 times the average maximum width detected by the views on the passage thus far). While this last condition eliminates exploration of most rooms, it remains a heuristic, as demonstrated by the three rooms circled in Figure \ref{fig:highwayexample}.

\begin{table}[t]
\caption{Features captured at each decision point}
\centering
\begin{tabular}{p{3in}}
\hline
Distance in front of the robot (average, maximum and minimum) \\
Distance to the robot's left (average and maximum)\\
Distance to the robot's right (average and maximum)\\
Full view of all distances (average, maximum, minimum, median, and standard deviation) \\
\hline
\end{tabular}
\label{tab:features}
\end{table}

While passages are likely to be more than $1m$ wide, real-world walls are unlikely to align neatly with any grid's arbitrary cell borders. HLC uses \textit{occupancy mapping} to add qualified cells on the robot's immediate left and right to a passage. This technique distinguishes between distance readings that extend beyond a neighboring cell (\textit{passes}) and readings that end within the cell (\textit{hits}). For a cell with $p$ passes and $h$ hits in a given view, when $h/(h+p) \leq 0.5$ the cell is termed \textit{unobstructed}; otherwise it is \textit{obstructed}. At each decision point, HLC adds a cell to a passage only if the cell is unobstructed in the current view and adjacent (within $2m$ of the robot and to its immediate right or left). Such a cell is labeled with the current passage number in the passage grid. Obstructed cells within $4m$ are labeled as such in the passage grid. Occupancy mapping results in the jagged nature of the passages in Figure \ref{fig:highwayexample}, where some space beyond open doorways is recorded as unobstructed and some cells near obstacles as obstructed.

\subsection{The passage network and the skeleton}
Passages produce connectivity only when they link to one another. Thus, as it extends a candidate, HLC also builds a \textit{passage network}, a graph where each vertex represents a labeled passage grid cell. The network records an edge between two vertices if the robot ever travels between the grid cells successfully, or if some view ever labeled an adjacent cell as unobstructed. HLC grows and uses this network while it explores for passages.

Once HLC completes exploration of a candidate, it removes candidates one at a time from the top of the candidate list until it finds one whose stretch remains uncovered in the passage grid and is dissimilar from any already explored candidate. HLC assigns that candidate a new passage number, but then must travel to the candidate's start and orient the robot in the direction of the stretch. If the robot is near (within $1 m$) the new start, HLC moves to within $0.5 m$ of it; otherwise, the robot must find its way to the new start.

Before the robot returns to a new candidate's start, the robot is already in some passage and the candidate was first glimpsed during exploration along some (possibly different) passage. In that case, HLC finds the shortest path in the passage network from the robot's current location to the candidate's start with breadth-first search. (The heuristic formation of the network does not support an appropriate underestimate for use with A*.) To transform that sequence of passage grid cells into a path, HLC replaces each cell with two decision points: the one closest to the center of that cell and the one closest to the center of the next cell in the sequence. This path is guaranteed to lead to the new candidate's start. Exploration of the world continues until no candidates remain or a specified time has elapsed. If the candidate list becomes empty, HLC makes one last attempt to find additional candidates --- it travels to the labeled cell farthest from its current position in the passage grid.

\begin{figure}[t]
\centering
\includegraphics[width=\linewidth]{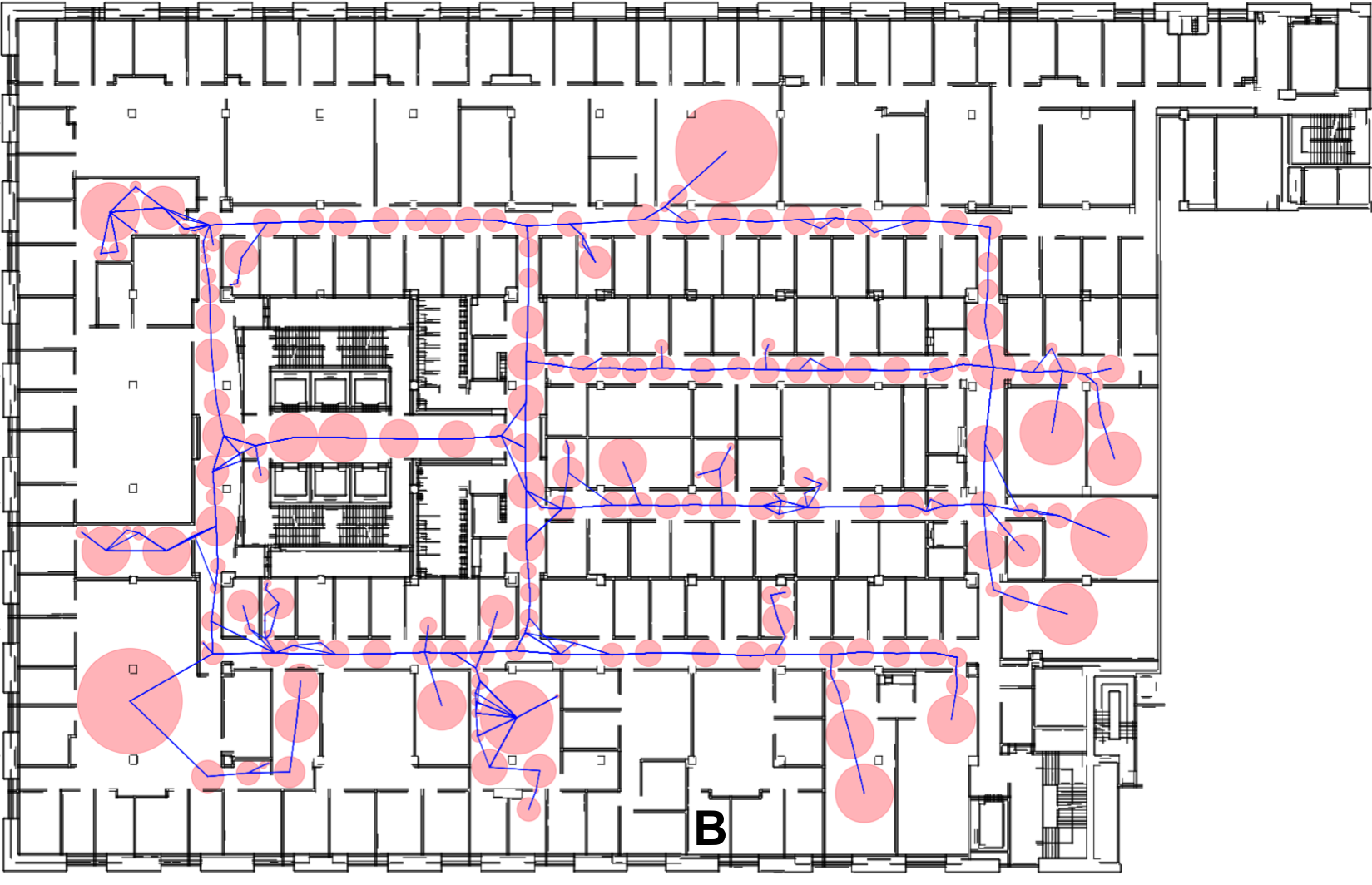}
\caption{Regions (circles) and edges form a skeleton.}
\label{fig:skeletonexample}
\end{figure}

HLC also builds a \textit{skeleton}, a graph of connected, unobstructed areas that supports long-range planning. After exploration, only the skeleton is preserved to support path planning. The skeleton re-represents each passage as a chain of \textit{regions}, non-overlapping, circular areas built from the decision points during the passage's exploration. An edge in the skeleton indicates that the robot moved directly from one region to the other without passing through any intervening region. The shortest of those transitions produces the label for that edge: its metric distance and the two \textit{endpoints} where the robot crossed the circumference of each region. Figure \ref{fig:skeletonexample} shows an example of a learned skeleton in GCWorld. Learning during subsequent travel continues to update the skeleton. 

\begin{figure}[t]
\centering
\includegraphics[width=\linewidth]{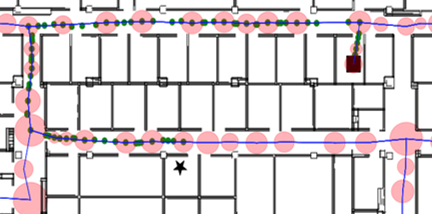}
\caption{In part of a skeleton, waypoints form a plan from the robot (square) to its target (star).}
\label{fig:planexample}
\end{figure}

\section{Planning and Navigation}
To contend with unanticipated and long-term challenges, people combine their learned spatial model with decision-making heuristics that balance reactivity and planning \cite{tenbrink2006wayfinding}. SemaFORR also integrates planning with reactive heuristics. A \textit{plan} here is a sequence of \textit{waypoints}, intermediate subgoals on the way to a target. Given a target, SemaFORR's high-level reasoning structure makes a plan to reach it, and tries to execute that plan, guided by two rules: "go to the next unobstructed, unvisited waypoint" and "avoid moves into obstructed locations." 

Ideally, SemaFORR would simply go from one waypoint in its plan to the next. The next waypoint, however, may be occluded or out of range, more than one action may adequately address it, or the robot's sensors or actuators may err. Because exploration captures only limited, passage-oriented data, the robot may also not be near the first region in its skeleton-based plan and the target may not be near the last one. In all these cases, SemaFORR relies on the combined opinions of its heuristics, which capture commonsense (e.g., ''move in the direction of the target'') and model-based reasoning (e.g., ''leave a dead-end that does not contain the target''). (See \cite{epstein2019planning} for additional details.) 

Previously, when given a target, SemaFORR applied A* search to a graph built from a fine-grained grid superimposed on a map of the world, with edge costs modified by its spatial model. In large worlds, however, the model's coverage of the world was determined by the targets set for the robot. HLC changes the way SemaFORR plans. 

With HLC, SemaFORR now deliberately explores for high-level connectivity first, and then plans in the (far smaller) skeleton instead of in a detailed metric map. The planner uses A* on the skeleton to find the shortest path between the robot and the target location. If the target is in a region, it uses that region as a vertex. In the more likely case where the target is outside a region, the planner first tries to find a region with \textit{visibility}, where a view associated with that region is able to detect the target. If no region has visibility, then the planner scores each region with $s = -5 * distance + degree$, where \textit{distance} is the Euclidean distance from the target to the region's center and \textit{degree} is the region's degree in the skeleton. This expresses a preference for regions that are not isolated and yet close by. The planner then designates the region with the maximum score as the endpoint for the plan. If the robot is not in a region, a similar process selects some region as the first point in the plan.

The resultant partial plan is a sequence of waypoints that interleave the regions' centers with their connecting edges' endpoints and the midpoint on the path between the regions. Figure \ref{fig:planexample} shows a sequence of waypoints in part of a skeleton. The robot must rely on voting among its heuristics for the initial and final legs of its travel.

\section{Experimental Design and Results}
To evaluate the impact of exploration on navigation, we conducted multiple simulation experiments in GCWorld with Freight, a Fetch Robotics robot. GCWorld's many jogs, interior posts, and narrow doorways, provide substantial challenges for Freight's $0.8m$-wide body. Fifteen times per second, the view from Freight's laser range finder measures 660 distances to the nearest obstruction within $25m$ (every 1/3\degree \ along a 220\degree \ arc). SemaFORR selects each action; the simulator executes it and reports the resultant pose and view. 

Freight's \textit{task} is to visit a preselected ordered list of 40 randomly chosen targets. An \textit{experiment} repeats each of 5 different tasks 5 times, for a total of 25 \textit{runs}. Freight \textit{fails} on a target if it does not reach it within 750 actions. The robot begins every run in the same pose, and begins on each target after the first from its final pose on the previous one. 

This experiment was executed with exploration (SemaFORR) and with \textit{SemaFORR-A}, an ablated version that gradually constructs the model without an initial exploration phase. Performance metrics, averaged over 25 runs, were total travel time in seconds, total travel distance in meters, percentage of targets reached successfully, and \textit{coverage}, the fraction of the world's unobstructed footprint covered by the spatial model, as evaluated in a $1m \times 1m$ grid.

Parameter values were based on preliminary analysis. HLC's exploration was limited to 750 decisions per candidate and 20 minutes, with moves of $0.8m$ to match Freight's size. Freight moved away from any obstruction within $0.15m$. A stretch had to be at least $7m$, and passage orientation was averaged over the last 40 decision points. Statistically significant differences appear in boldface ($p=0.05$). 

\begin{table}[t]
\caption{Success and Coverage}
\centering
\begin{tabular}{lrrr}
\hline
Navigator  & Success Rate & Initial Coverage & Final Coverage\\ \hline
SemaFORR-A   & 46.17\%      & 0.00\% & 22.97\%\\ 
SemaFORR    & \textbf{70.60\%}   & 12.65\%  & \textbf{27.94\%} \\
\end{tabular}
\label{tab:resultssuccess}
\end{table}

SemaFORR was significantly more successful than SemaFORR-A in this experiment. (See Table \ref{tab:resultssuccess}.) Table \ref{tab:resultstravel} compares their travel time and distance, and breaks out the task-only data from SemaFORR. Even when exploration is included, SemaFORR required significantly less time than SemaFORR-A. Although SemaFORR explored for an average of 1084.10 seconds, that initial effort saved more than twice as much time when it went on to address the tasks. 

Travel distance for the tasks was also significantly shorter. Even when exploration distance is included, SemaFORR did not travel significantly further than SemaFORR-A. SemaFORR-A made 63.80\% of its decisions with heuristics, while SemaFORR, whose plans were more reliable, resorted to heuristics significantly less often (49\%).

\begin{figure}[t]
\centering
\includegraphics[width = \linewidth]{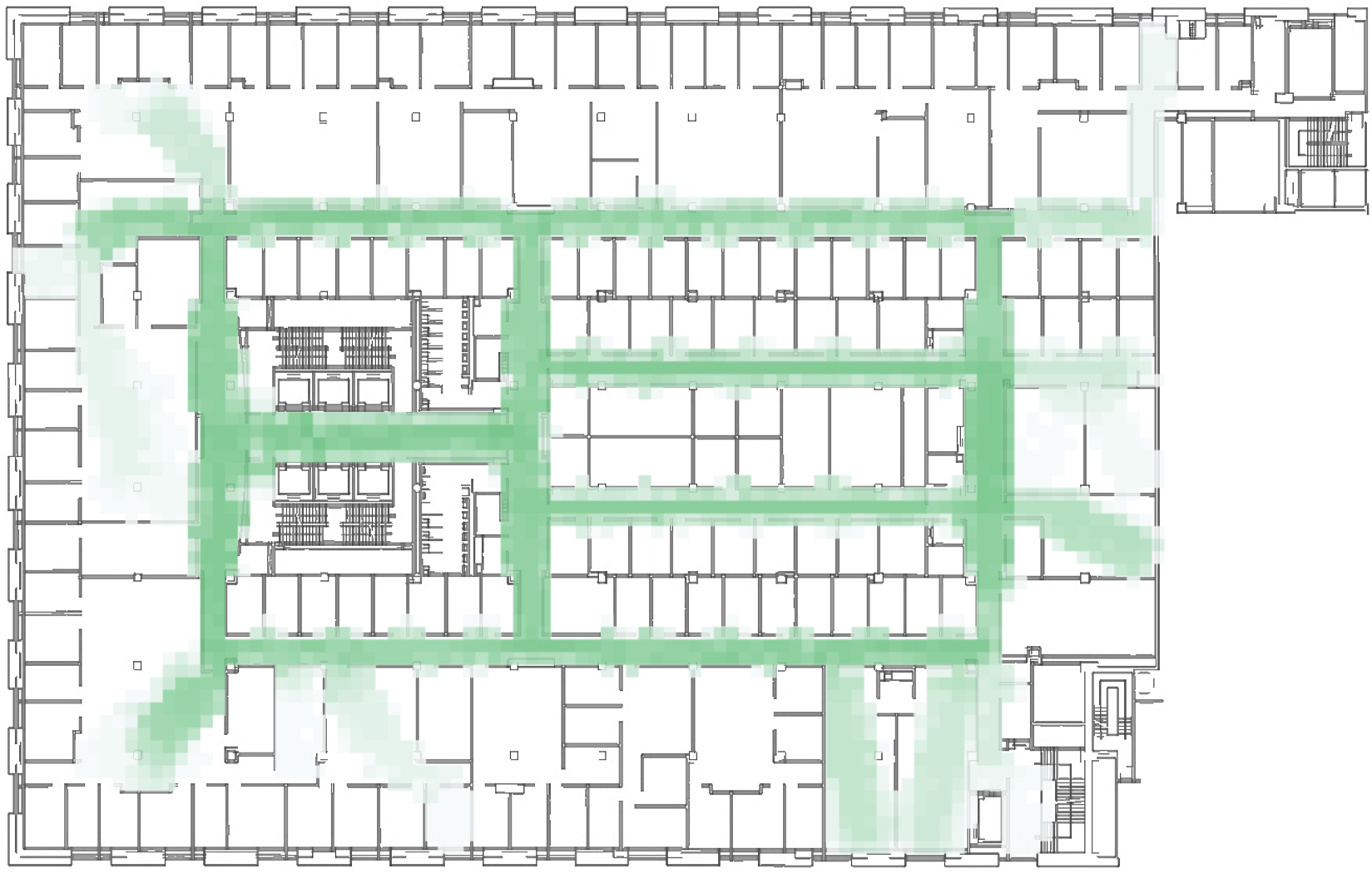}
\caption{Average passage coverage over 25 runs in GCWorld. Darker cells were detected more often.}
\label{fig:averagehighwaycoverage}
\end{figure}

SemaFORR's coverage of GCWorld was 12.65\% after exploration, and twice that after it had attempted all its targets. Figure \ref{fig:averagehighwaycoverage} shows the frequency with which HLC captured each passage grid cell. Clearly it found GCWorld's principal hallways and avoided most other areas. On average, HLC covered 82.88\% of the 8 readily recognized hallways in Figure \ref{fig:GCWorld}. Table \ref{tab:resultssuccess} shows that SemaFORR-A's final model covers less of GCWorld than SemaFORR's does. Indeed, although the two systems covered about as much of GCWorld with their model, SemaFORR's model was clearly more effective. 

\begin{table}[t]
\caption{Travel Time and Distance}
\centering
\begin{tabular}{lrr}
\hline
Navigator  & Travel Time & Distance \\ \hline
SemaFORR-A   & 8216.45  & 5889.49         \\ 
SemaFORR (tasks only)    & \textbf{5608.93}      & \textbf{4610.87}          \\
SemaFORR (tasks and exploration)    & \textbf{6693.02}      & 5769.50           \\
\end{tabular}
\label{tab:resultstravel}
\end{table}

SemaFORR's planning in the skeleton averaged 3.56 seconds per task. In previous work, we showed that A* planning on the GCWorld grid for similar tasks averaged about 15 seconds, and planning with a model-based A* graph averaged about 1 minute. This speedup occurs because the scale of the skeleton graph (205 vertices and 248 edges on average at the end of a task) is several magnitudes smaller than the grid-based graph (85,000 vertices and 170,000 edges). The skeleton captures a far coarser representation of the world and yet does not require the metric details of an A* graph to produce relatively successful travel.

\section{Discussion}
Human skill at navigation in novel worlds inspired this work. Although an early version of SemaFORR proved successful in a variety of worlds like Figure \ref{fig:laserscan}, it had difficulty as they became more complex. 
Moreover, embodiment with an industrial-sized robot in far larger, more realistic spaces like GCWorld repeatedly stymied that version with its complexity and the deceptive proximity of targets just beyond some nearby wall. SemaFORR seemed to need a map in which to plan. Instead we augmented it with self-guided exploration.

Given Figure \ref{fig:highwayexample} or a road map, most people and efficient, direction-giving algorithms quickly notice lengthy, readily navigable spaces like our passages. They identify nearby ones and then search to reach them from their current location and from the target. SemaFORR shows that even limited exploration provides a sufficiently robust model for planning without a map. We have tested this premise successfully on other, differently configured floors in the same building as GCWorld, and are currently examining it in a building whose footprint is not rectangular. 

SemaFORR continues to modify its skeleton after exploration, when it visits targets. As a result, the original ”pure” passage network becomes embellished with chains of regions that represent shorter stretches or dead-ends (rooms). Although planning may become less efficient as the skeleton grows, the expanded cognitive model also supports more informed decisions and planning within previously unvisited areas. Prioritization of the regions that originated within passages is a subject of current work. 

We considered planning in a labeled topological graph where nodes represented passages and edges represented their intersections. That graph would be even smaller than the skeleton. An intersection there would afford extensive travel, and thus be a place of interest and a potential landmark. It would require considerable computation, however, to construct this graph from the passage grid, which is not well-aligned with obstructions in the world, has cells larger than many regions, and must retain the full decision history in memory. A better approach, currently under development, is to coalesce the regions in the obstruction-free, more memory-efficient skeleton to produce a representation more like human cognitive spatial models. 

When SemaFORR failed on a task here, it was usually because the skeleton or the spatial model lacked sufficient detail in a relevant area. For example, the adjacent but poorly connected rooms nearby to the room labeled B in Figure \ref{fig:skeletonexample} make navigation particularly difficult. Such rooms are analogous to a side street off of a main road off a highway, where knowledge of the highway network is not enough to get to that side street. As the skeleton grows with experience, however, SemaFORR is able to succeed on such failed targets upon new attempts. (About 62.5\% of initial failures averaged over three runs were successful after repeated attempts.)

There are several caveats with our approach. Only exhaustive exploration can guarantee perfect knowledge of an environment. Without it, navigation and plans for it can be less than ideal. (For example, a shorter path in Figure \ref{fig:planexample} goes undetected because of a gap in the skeleton.) Moreover, HLC can only learn passages if can find at least one stretch to explore. Like architects, we had our robot enter its world for the first time where it would detect a stretch (i.e., at the elevators). Search for a first stretch instead is future work, as is extension to three dimensions. In addition, we assumed near-perfect localization, that is, that the robot knew exactly where it and the target were. In built spaces this can be achieved with additional sensors.

Finally, the cognitive spatial model described here has other uses. It is a descriptive device for designers of indoor spaces, and could be predictive for floors of the same building. It can also be used to generate user-friendly explanations even when a robot is governed by another controller. Assuming you begin facing to the right in Figure \ref{fig:GCWorld}, for example, a solution could be described at a high level as ''go to the end of the hallway, turn left, go to the end of that hallway, turn right, and your destination will be just before the end of that hallway, on your left." Meanwhile, limited exploration provides a robust model for navigation in unfamiliar worlds.


\addtolength{\textheight}{-12cm}   








\bibliographystyle{IEEEtran}
\bibliography{IEEEabrv,IEEEexample,references}

\end{document}